\RequirePackage{fix-cm}
\documentclass[smallcondensed]{svjour3}     
\smartqed  
\usepackage{graphicx}
\graphicspath{{figures/}}
\usepackage{tikz}
\usepackage{pgfplots}
\pgfplotsset{width=8.8cm, 
            height=6cm, 
            compat=1.9, 
            every axis/.append style={font=\footnotesize},
            }
\usepackage{verbatim}
\usepackage{textcomp}
\usetikzlibrary{shapes,arrows}
\usepackage{amsmath} 
\usepackage{amssymb}  

\usepackage{algorithm}
\usepackage[noend]{algpseudocode}
\renewcommand{\algorithmicrequire}{\textbf{Input\;\:\::}}
\renewcommand{\algorithmicensure}{\textbf{Output:}}

\usepackage{color}

\begin{document}

\title{A visual study of ICP variants for Lidar Odometry
}


\author{Sebastian Dingler         \and
        Hannes Burrichter 
}


\institute{Daimler Truck AG \at
              70546 Stuttgart \\
	      Germany \\
           ORCiD: 0000-0002-0162-8428 \\
              \email{sebastian.dingler@daimler.com} 
           \and
           Technical University of Munich \at
           80333 Munich \\
	   Germany \\
           \email{hannes.burrichter@tum.de}
}

\date{Submitted to JINT 06.01.2021}

\maketitle

\begin{abstract}
Odometry with lidar sensors is a state-of-the-art method to estimate the ego pose of a moving vehicle.
Many implementations of lidar odometry use variants of the Iterative Closest Point (ICP) algorithm.
Real-world effects such as dynamic objects, non-overlapping areas, and sensor noise diminish the accuracy of ICP.
We build on a recently proposed method that makes these effects visible by visualizing the multidimensional objective function of ICP in two dimensions.
We use this method to study different ICP variants in the context of lidar odometry.
In addition, we propose a novel method to filter out dynamic objects and to address the ego blind spot problem.
\end{abstract}

\section{Introduction}
\label{intro}
Odometry is a widely used technique to estimate the trajectory of a navigating robot or an automated driving system (ADS). 
As more and more ADSs make their way to higher automation levels, the demand for higher accuracy and robustness rises accordingly. 
One example of this demand is the use of high-definition maps in such systems. 
To leverage the map information, the AD system first needs to localize itself in the map. 
Therefore, the accuracy of the localization influences the overall confidence that can be placed in the map information. 
Since odometry plays an essential role in many localization systems, the robotics community is interested in improving its accuracy and robustness. 

In recent years, active sensors such as lidars (Light Detection and Ranging) have become very popular. 
Lidars have the advantage that they produce rich 3D point clouds with high frequency and a wide field of view. 
At the same time, the error of the range measurements is nearly constant irrespective of the distance. 
For this reason, lidar sensors are one of the most widely used sensor modalities for automated driving systems.
Lidar odometry, in its most common form, matches a current lidar scan with a previously acquired scan. 
By concatenating all scan-to-scan estimates, the ego trajectory can be estimated. 

For the scan matching, variations of the Iterative Closest Point (ICP) algorithm are used. 
ICP estimates rotation and translation iteratively. 
Unfortunately, ICP has some difficulties such as local minima and sensitivity to noise and outliers. 
Furthermore, in real-world applications, each scan samples the environment differently, which leads to point pairs without meaningful correspondences. 

Since the original concept of ICP only works well with ideal data, hundreds of variations have been proposed \cite{Pomerleau2013}. 
Even though comparisons exist, from our experience it is still necessary to reimplement several variants and to evaluate which variant fits best for lidar odometry. 
One reason is that studies like Rusinkiewicz and Levoy's focus on the general registration problem and not on lidar odometry \cite{Rusinkiewicz2001}. 
This motivated us to conduct a study specifically for lidar odometry by visualizing the objective function of several ICP variants.

This paper is an extended version of our conference paper by Dingler and Burrichter with additional ICP variants and newer experiments \cite{Dingler2019}.
To be more specific, since we investigate additional variants of ICP, we structured the ICP chain inspired by the work of Pomerleau et al. \cite{Pomerleau2013} and Rusinkiewicz et al. \cite{Rusinkiewicz2001}.
Our structure of the ICP process contains the following stages: 
\begin{enumerate}
    \item Data filtering 
    \item Data processing
    \item Correspondence determination
    \item Objective function
    \item Minimizer
\end{enumerate}
Except for the last stage, we study additional variants compared to our conference paper. 
For example, we investigate the Lidar Odometry and Mapping (LOAM) method, which is the top-ranking lidar-odometry approach among the lidar-only methods on the KITTI odometry benchmark dataset \cite{Zhang2014}. 
Furthermore, this paper proposes two novel data filtering and data processing approaches,
namely the ego blind spot filter and the Octree Correspondence Filter (OCF). 
The first addresses a blind spot that is present in many sensor configurations where the field of view of the lidar does not cover the close vicinity of the ego vehicle.
This blind spot causes non-overlapping areas in the point cloud, which later disturb the alignment process.
The Octree Correspondence Filter is able to filter out wrong point correspondences that are caused by dynamic objects and non-overlapping areas.

Previous work that relates to ours includes authors who have analyzed the convergence of ICP.
For example, Mitra et al. present a figure that shows the funnel of convergence of the point-to-plane objective function \cite{Mitra2004}. 
They perform ICP with many different initial transformations and visualize if ICP converges or not.
Magnusson et al. conducted similar experiments with real-world data \cite{Magnusson2009}.
The most closely related work to ours is Tazir et al. \cite{Tazir2018}.
The authors plot the objective function in two dimensions by sampling it for translational and rotational couples.
For each couple, they analyze the convexity of the objective function in a qualitative manner.
The recent paper from Landry et al. visualizes the point-to-plane objective function by projecting it to an $\mathbb{R}^2$ plane \cite{Landry2019}.
The purpose is to show that the objective function is smooth at a larger scale and around the ground truth, it becomes rough due to reassociations. 
All works have in common that the visualization is exclusive to one or two dimensions of the domain of the objective function. 
Our method instead is able to cover all dimensions of the domain but is less flexible in exploring it.

Beyond the scope of this paper are probabilistic ICP variants like Generalized-ICP (G-ICP) \cite{Segal2009} and Normal Distributions Transform (NDT) \cite{Magnusson2009b}.
Even though we regard them as compelling alternatives to the standard ICP approaches, we do not examine them due to their conceptual differences. 
In addition, we focus only on automotive lidar odometry because of the limited spatial movement capabilities of an on-road vehicle.

The rest of the paper is organized as follows. 
The next section summarizes ICP and the major variants and introduces our notation. 
In Section \ref{visualization_method}, we present our visualization method. 
Finally, Section \ref{experiments} studies ICP variants based on our visualization method and we make our conclusion in Section \ref{conclusion}.

\section{ICP and its Variants}
\label{icp_variants}
The Iterative Closest Point or Iterative Corresponding Point (ICP) algorithm is a widely used method to align two point clouds. 
The alignment of the point clouds is achieved by identifying corresponding point pairs and iteratively minimizing the distance between those pairs. This fundamental concept has not changed since the first introduction of ICP by Chen and Medioni \cite{Chen1991} and Besl and McKay \cite{Besl1992}. 
Nonetheless, many researchers have proposed improvements to ICP since the original approach has some limitations in real-world applications. 
We will present some of these improvements in the context of lidar odometry. 
First, we formalize the general alignment problem, and then we state the ICP variants we are studying.

\begin{figure}
\tikzset{%
  block/.style    = {draw, thick, rectangle, minimum height = 3em,
    minimum width = 3em},
  block2/.style    = {draw, dashed, rectangle, minimum height = 3em,
    minimum width = 3em},
  input/.style    = {coordinate, node distance = 1cm}, 
  output/.style   = {coordinate, node distance = 1cm} 
}
\newcommand{\suma}{\Large$+$}
\newcommand{\inte}{$\displaystyle \int$}
\newcommand{\derv}{\huge$\frac{d}{dt}$}
\begin{tikzpicture}[auto, thick, node distance=2cm, >=triangle 45]
\draw
	node [input, name=input1] {}
	node at (1.5,0)[block, align=center] (datafiltering1) {Data \\ filtering}
	node at (3.5,0)[block, align=center] (dataprocessing) {Data \\ processing}
        node at (6,0)[block, align=center] (correspondence) {Correspondence \\ determination}
        node at (8.5,0)[block, align=center] (objective) {Objective \\ function}
	node [block2, right of=objective] (minimizer) {Minimizer}
        node at (12.2,0)[output] (sal2){};

	\draw[->](input1) -- node {$\mathcal{P}, \mathcal{Q}$}(datafiltering1);
        \draw[->](datafiltering1) -- node {}(dataprocessing);
 	\draw[->](dataprocessing) -- node {}(correspondence);
        \draw[->](correspondence) -- node {}(objective);
        \draw[->](objective) -- node {}(minimizer);
        \draw[->,dashed](minimizer) -- node {}(sal2);
        \draw[->,dashed](minimizer) -- (11.7,0) |- (3.8,-1.5) -- (3.8,-0.5) node {}(dataprocessing);

\draw
	node at (11.7,0) {\textbullet}; 

\draw
	node at (3.2,-1.7)[name=input2]{$\mathbf{T}_u$};

\draw
	node at (9.4,0.3)[name=input2]{$\epsilon$};
\draw
	node at (11.7,0.3)[name=input2]{$\mathbf{T}_e$};

\draw[->](input2) (3.2,-1.5) -- (3.2,-0.5) node {}(dataprocessing);

\end{tikzpicture}
        \caption{Simplified architecture of ICP with five processing stages. The data filtering stage subsumes functionalities that are done offline, thus before starting the iteration, e.g. downsampling of the point clouds. The data processing stage works online with the point clouds. At least, it applies the current estimate of $\mathbf{T}_e$ to $\mathcal{P}$. The correspondence determination stage, establishes the corresponding point pairs $(p_i \in \mathcal{P}, q_j \in \mathcal{Q})$. Subsequently, the point pairs are evaluated with an objective function $\epsilon$ which will be minimized. The minimizer passes its current estimate $\mathbf{T}_e$ back to the data processing stage. However, our visual method can be seen as an open-loop study of ICP. Instead of passing the current estimated transformation $\mathbf{T}_e$ from the minimizer, we pass interpolated transformations $\mathbf{T}_u$ to control the behavior of the building blocks and visualize the objective function $\epsilon$.}
\label{fig:icp_loop}
\end{figure}
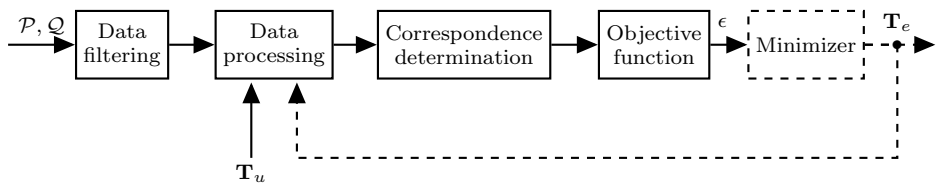

Let $\mathcal{P}$ and $\mathcal{Q}$ be two point clouds in the $d$-dimensional Euclidean space $\mathbb{R}^d$. 
We indicate a point within $\mathcal{P}$ with lowercase character $p$ and a point in $\mathcal{Q}$ with $q$. 
The intent of ICP is to estimate the relative transformation that, when applied to $\mathcal{P}$, aligns both point clouds into a common coordinate system. In three-dimensional space, the transformation 
\begin{equation}
\mathbf{T}  = 
\left\{ \left( \begin{array}{lc}
\mathbf{R} & \mathbf{t} \\
\mathbf{0}_{1 \times 3} & 1 \\
\end{array} \right) \;\bigg\vert\; \mathbf{R} \in SO(3) \mbox{ and } \mathbf{t} \in {\mathbb{R}}^3 \right\}
\end{equation}
lies in the special Euclidean group $SE(3)$ where we later refer to $\mathbf{R}$ as the rotation matrix from the special orthogonal group $SO(3)$ and to $\mathbf{t}$ as the translation vector. 
Given an initial estimate $\mathbf{T}_0$, the key idea of ICP in its simplest form is to determine corresponding point pairs $(p_i \in \mathcal{P}, q_j \in \mathcal{Q})$ and to minimize an objective function $\epsilon$ to estimate a new transformation $\mathbf{T}_1$.
The classical correspondence search is done by finding the closest point for all points of cloud $\mathcal{P}$ in cloud $\mathcal{Q}$:
\begin{equation}
    p_i \mapsto q_j \quad \text{for } \; \forall p_i \in \mathcal{P}
\end{equation}
where $ p \mapsto q $ denotes the correspondence of $p$ to $q$.
Typically, the closest point search is implemented with kd-trees which leads to relatively good querying speed \cite{Bentley1975}.
The correspondence search is repeated in every ICP iteration because the correspondences are affected by the current transformation estimate. 
Once the objective function converges and falls below a given threshold, the alignment is complete. 
One fundamental challenge is to find the right correspondences that facilitate the estimation process. 
Ideally, the correspondence should be between points that are the closest in position when the two point clouds are aligned perfectly. 
If the correspondences are incorrect, ICP runs the risk of converging to a local minimum.
In many real-world applications, correct correspondences are unknown or simply not present due to multiple real-world problems.
We categorize them in the following way:
\begin{enumerate}
    \item Partial non-overlapping scans due to different scanning locations can lead to differences in the scanned objects of both point clouds resulting in points without any correct correspondence. \label{Correspondence problem 1}
    \item The presence of dynamic objects leads to different locations of the same object in consecutive scans. \label{Correspondence problem 2}
    \item Due to finite sensor resolution, measurements never correspond to the same location of the physical surface. Therefore, the objective function can never reach a value of zero even with perfect alignment. \label{Correspondence problem 3}
\end{enumerate}
To account for these problems, researchers have come up with more robust strategies. 
To keep track of the ICP variants, we have structured ICP into five processing stages. 
Figure \ref{fig:icp_loop} outlines these stages.
Next, we will introduce at least one variant per stage, starting with different objective functions.

\subsection{Point-to-point}
\label{sec:point-to-point}
In its simplest form, the ICP algorithm uses the point-to-point metric. The objective function
\begin{equation}
    \epsilon_{\text{point-to-point}}(\mathbf{T}) = \sum^n_{i=1} \left\| \mathbf{T}(p_i) - q_j \right\|^{2}
\end{equation} 
where $p_i \mapsto q_j$ denotes the correspondence between $p_i$ and $q_j$ and the metric measures the squared Euclidean distance between these points. 
Besl and McKay have shown that the ICP algorithm always converges monotonically to a local minimum with respect to this distance \cite{Besl1992}. 

\subsection{Point-to-plane}
\label{sec:point-to-plane}
The widely used point-to-plane error metric, originally introduced by Chen and Medioni \cite{Chen1991}, allows the generation of a real point to a virtual point correspondence. 
The error metric calculates the alignment error between a point $ p_i $ and the sub-space spanned by the surface normal $n_{q_j} $ of the corresponding point $ q_j $ which can be formulated as
\begin{equation}
    \epsilon_{\text{point-to-plane}}(\mathbf{T}) = \sum^n_{i=1} \left[ (\mathbf{T}(p_i) - q_j) \cdot n_{q_j} \right]^{2} 
\end{equation}
with $p_i \mapsto q_j$ denoting the correspondence.
This requires the calculation of the normals $n_{q_j}$ of $\mathcal{Q}$, which is usually done by using neighboring points of $q_j$. 
Unfortunately, this preprocessing step can introduce numerical errors. 
However, this procedure allows $p_i$ to slide along the plane spanned by $q_j$.
This can mitigate problem \ref{Correspondence problem 3}) as the sampling has no further influence on the alignment error for planar surfaces.


\subsection{Symmetric objective function}
\label{sec:symmetric}
Rusinkiewicz \cite{Rusinkiewicz2019} recently improved the point-to-plane objective function by proposing a symmetrized version.
The idea is to increase the freedom of movement for a greater class of surfaces instead of just planar ones. For that reason, he uses the normal vectors $n_{p} $ and $n_{q} $ of both corresponding points. The symmetric objective function is
\begin{equation}
    \epsilon_{\text{symm}}(\mathbf{T}) = \sum^n_{i=1} \left[ (\mathbf{T}(p_i) - q_j) \cdot (n_{p_i} + n_{q_j}) \right]^{2} 
\end{equation}
with $p_i \mapsto q_j$ denoting the correspondence. 
Note that the transformation $\mathbf{T}$ is not applied to $n_{p_i}$ and we can confirm this simplification that Rusinkiewicz has explored.
Like the point-to-plane metric, the symmetric objective function addresses the influence of the sampling problem \ref{Correspondence problem 3}).

\subsection{LOAM-Features}
\label{sec:loamFeatures}
The Lidar Odometry and Mapping (LOAM) method is a two-step lidar-odometry approach with (1) a high-frequency and low-accuracy stage for a rough frame-to-frame alignment and (2) a low-frequency and high-accuracy frame-to-map alignment stage \cite{Zhang2014}. 
Furthermore, the alignment of the frames is done by minimizing two distance metrics at the same time using feature points. 
Namely, edge point to edge line and planar point to planar patch. 
The computation of the edge points and planar points varies depending on the processing stage of LOAM. 
For example, in the first LOAM paper, a spinning 2D lidar was used and the edge lines and planar patches were determined based on knowledge of how the points were acquired.

To conduct a fair comparison with the other ICP variants, we evaluate LOAM in the following way. 
First, we analyze the two distance metrics independently from each other. 
Second, we use the distance metrics from the frame-to-mapping stage because this step is more independent of the sensor setup and aims for higher accuracy. 
Since we are only interested in frame-to-frame alignment, we set the number of frames for the map to one. 
Through this we investigate the edge point to edge line distance, and the planar point to planar patch distance. 

Zhang and Singh determine the edge points and planar points by computing a smoothness value $s_i$. For each point $p_i \in \mathcal{P}$ a set $\mathcal{S}_i$ of surrounding points is chosen. For that set the smoothness value 
\begin{equation}
s_i = \frac{1}{|\mathcal{S}_i| \cdot || p_i ||} || \sum_{p_j \in \mathcal{S}_i}  (p_i - p_j) ||
\end{equation}
is computed where $|\mathcal{S}_i|$ denotes the cardinality of the set $\mathcal{S}_i$. 
We note that $|| p_i ||$ introduces distance weighting.
Zhang and Singh define edge points with large $s_i$ values and planar points with small $s_i$ values.
Based on the smoothness value, edge points and planar points can be computed, which will be used in the following metrics.

\subsubsection{Edge point to edge line}
\label{sec:edgePointsToEdgeLines}
If $p_i$ is classified as an edge point, an edge line needs to be found in $\mathcal{Q}$. For this purpose, a set $\mathcal{Q}_i \subset \mathcal{Q}$ of the closest points to $p_i$ is composed. Then, the eigenvalues and eigenvectors are computed for $\mathcal{Q}_i$. If the eigenvalues consist of one large and two small eigenvalues, an edge line is found as the correspondence to the edge point. 
With this correspondence, the following distance is formulated
\begin{equation}
\epsilon_{\text{edge-to-edgeLine}}(\mathbf{T}) = \sum^n_{i=1} \frac{||\left ( \mathbf{T}(p_i) - \hat{q}_j \right ) \times (\mathbf{T}(p_i) - \hat{q}_k)||}{||\hat{q}_j - \hat{q}_k||} \ .
\end{equation}
Please note, $\hat{q}_j$ and $\hat{q}_k$ are not real points of $\mathcal{Q}$, they are computed based on the eigenvector with the largest eigenvalue since this vector forms the edge line.

\subsubsection{Planar point to planar patch}
\label{sec:planarPointToPlanarPatch}
If $p_i$ is classified as a planar point, a corresponding planar patch in $\mathcal{Q}$ is searched for. 
Like for the edge points, a set $\mathcal{Q}_i \subset \mathcal{Q}$ of the closest points to $p_i$ is identified. 
Then, the eigenvalues and eigenvectors are computed for $\mathcal{Q}_i$. 
If the eigenvalues consist of two large and one small eigenvalue, the points of $\mathcal{Q}_i$ construct a planar patch.
The eigenvector that belongs to the smallest eigenvalue defines the orientation of the planar patch. 
With this correspondence, the planar point to planar patch distance is computed with
\begin{equation}
\epsilon_{\text{planar-to-planarPatch}}(\mathbf{T})  = \sum^n_{i=1}  \frac{\left | \left | \begin{array}{c}
\mathbf{T}(p_i) - \hat{q}_j\\
(\hat{q}_j - \hat{q}_k) \times (\hat{q}_j - \hat{q}_l) \\
\end{array}\right | \right |}{\left | \left |(\hat{q}_j - \hat{q}_k) \times (\hat{q}_j - \hat{q}_l) \right | \right |}
\end{equation}
where $\hat{q}_j$, $\hat{q}_k$ and $\hat{q}_l$ are computed points located on the planar patch.

\subsection{Reciprocal correspondence search}
\label{sec:reciprocal}
One approach to improve the correspondence search is to use reciprocal correspondences \cite{Pajdla1995}. 
A correspondence between points of two point clouds $\mathcal{P}$ and $\mathcal{Q}$ is considered reciprocal if both points $ p_i \in \mathcal{P}$ and $ q_j \in \mathcal{Q}$ share the same correspondence, formally
\begin{equation}
p_i \mapsto q_j \quad \text{and} \quad q_j \mapsto p_i \ .
\end{equation}
Reciprocal correspondence is suited to handle non-overlapping areas and dynamic objects as pointed out in problems \ref{Correspondence problem 1}) and \ref{Correspondence problem 2}).
To relax the strict reciprocal correspondence it is also possible to allow $q_j \mapsto p_k$, if $||p_i - p_k||$ is smaller than a given threshold.

\subsection{Ego overlap filter}
\label{sec:egoBlindSpot}
\begin{figure}[t]
    \centering
    \includegraphics[width=1.0\textwidth]{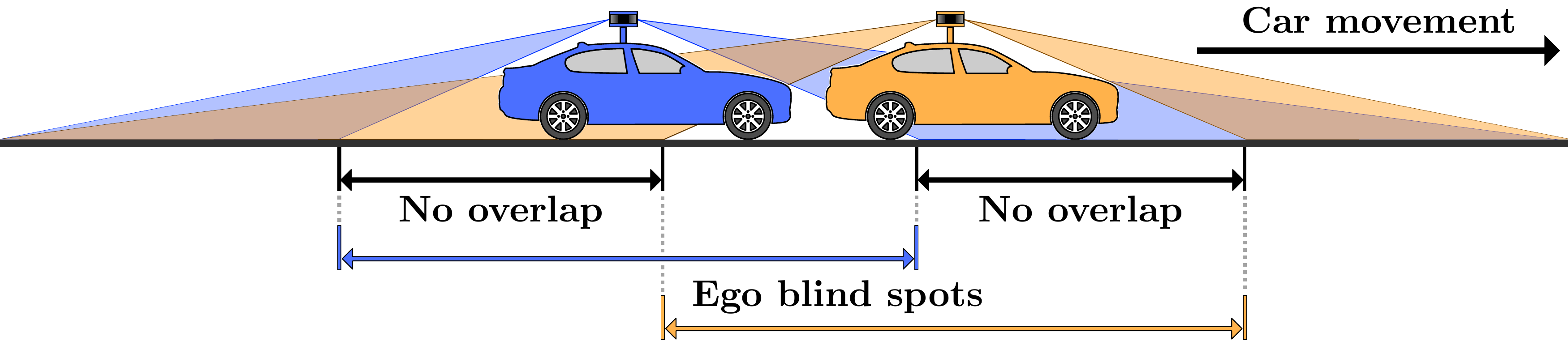}
    \caption{Illustration of the ego blind spots due to the sensor mounting position on the roof. The figure shows two consecutive positions of the same ego car moving towards the right. The ego blind spot creates non-overlapping areas in the two lidar scans, labeled as no overlap.}
    \label{fig:Ego Blind Spot}
\end{figure}

For lidar odometry, the most predominant mounting position of the sensor is the roof of the vehicle. 
This mounting position creates a circular blind spot in the vicinity of the vehicle that causes non-overlapping areas in the alignment process.
Non-overlapping areas are an example of problem \ref{Correspondence problem 1}) where the closest point search often establishes wrong correspondences.
Figure \ref{fig:Ego Blind Spot} illustrates this problem for a moving car. 

Since the size of the non-overlapping area depends on the temporal movement of the vehicle and the field of view of the lidar, we formulated an ego blind spot filter.
In its simplest form, the blind spot area can be modeled with a horizontal circle of radius $r$ in the $xy$-plane.
Given an estimate for the transformation between two poses, the points that are located within the ego blind spot can be filtered out because a corresponding point will not exist. 

Formally, if $[q_i]_x$ and $[q_i]_y$ are the $x$ and $y$ coordinates of $q_i \in \mathcal{Q}$ and $t_x$ and $t_y$ are the corresponding translations of the current estimated transformation $\mathbf{T}_e$, if
\begin{equation}
([q_i]_x - t_x)^2 + ([q_i]_y - t_y)^2 \leq r^2
\end{equation}
is true, the point $q_i$ will be in the ego blind spot and is therefore filtered out. 
For a point $p_i \in \mathcal{P}$ the same procedure holds, except the inverse $\mathbf{T}^{-1}_{e}$ of $\mathbf{T}_{e}$ is used.
The main challenge of this filtering approach is to find an appropriate transformation estimate $\mathbf{T}_{e}$.
For a moving car, the transformation estimate $\mathbf{T}_{e,n-1}$ from the previous frame can be used.

\subsection{Octree Correspondence Filter}
\label{sec:octreeFilter}
\begin{figure}
    \centering
\vspace{0.2cm}
    \includegraphics[width=1.0\textwidth]{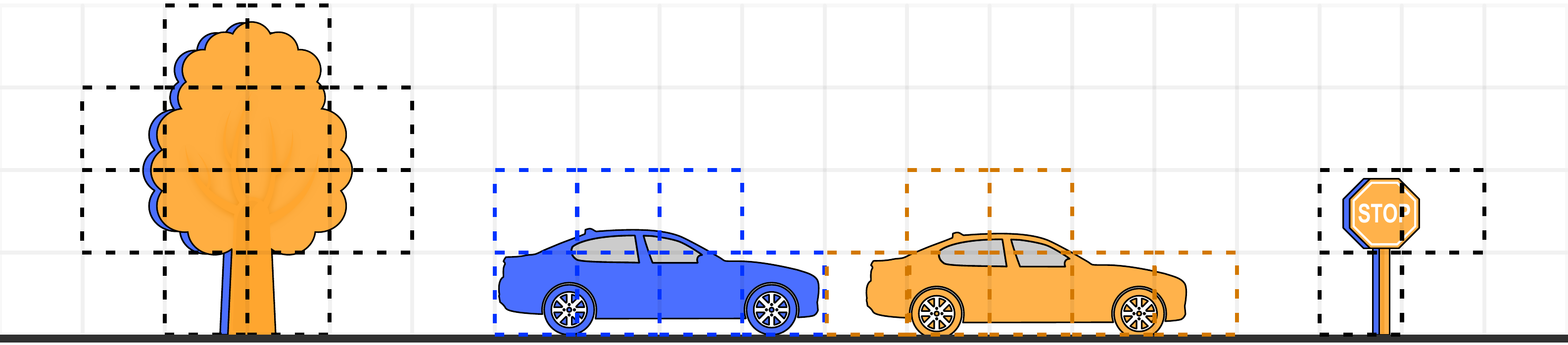}
    \caption{Illustration of the Octree Correspondence Filter after aligning frame $n$ and $n-1$, colored with orange and blue, in a common coordinate frame. The black squares denote voxels that are occupied in both scans. Blue and orange squares denote voxels only occupied by one of the two point clouds. Points within theses voxels will be deleted since they contain dynamic or non-overlapping points.}
    \label{fig:octreeFilter}
\end{figure}

This section proposes a novel algorithm to filter out points without correct correspondences, including dynamic objects and non-overlapping areas. 
The algorithm adopts the idea of splitting the point cloud into regions. One representative work of this idea is Bengtsson and Baerveldt \cite{Bengtsson1999}. 
They proposed splitting the point clouds into several large regions and matching each of them separately to the reference scan. 
The regions with a high mean correspondence error are then discarded. 
Our algorithm (listed in Algorithm \ref{alg:octree_corr_filter}) utilizes an octree data structure that recursively divides the three-dimensional space into eight cubic volumes called octants or voxels. 
With this data structure, a point cloud can be partitioned into addressable voxels with variable sizes.

 \begin{algorithm}
    \caption{Octree Correspondence Filter}
    \label{alg:octree_corr_filter}
    \begin{algorithmic}[1]
        \vspace{4pt}
        \Require Two point clouds: $\mathcal{P} = \{ p_i \}$, $\mathcal{Q} = \{ q_j \}$  \newline \phantom{\qquad\;\;} Transformation estimate: $\mathbf{T}_{e}$
        \vspace{2pt}
        \Ensure Filtered point clouds: $\mathcal{P}_{filtered}$, $\mathcal{Q}_{filtered}$
        \vspace{5pt}
        \Function{OctreeCorrespondenceFilter}{$\mathcal{P}$, $\mathcal{Q}$, $\mathbf{T}_{e}$}
        	\State $\mathcal{\tilde{P}} \;\;\, \gets \; \mathbf{T}_{e}(\mathcal{P})$
            \State $ O_\mathcal{\tilde{P}} \; \gets \; \Call{createOctree}{\mathcal{\tilde{P}}} $
            \State $ O_\mathcal{Q} \;\; \gets \; \Call{createOctree}{\mathcal{Q}\:} $
            \ForAll{voxel $ v_i \in O_{\mathcal{\tilde{P}}}, u_i \in O_\mathcal{Q} $}
                \If{$ \Call{isEmpty}{v_i} \, = true $ or $ \Call{isEmpty}{u_i} = true $}
                \State $ \; \Call{clearVoxel}{v_i} $
                \State $ \; \Call{clearVoxel}{u_i} $
                \EndIf
            \EndFor
            \vspace{2pt}
            \State $ \tilde{\mathcal{P}}_{filtered} \; \gets \; \Call{extractPoints}{O_\mathcal{\tilde{P}}} $
            \State $ \mathcal{Q}_{\:filtered} \; \gets \; \Call{extractPoints}{O_\mathcal{Q}} $
            \State $ \mathcal{P}_{filtered} \;\;\;\, \gets \; \mathbf{T}^{-1}_{e}(\tilde{\mathcal{P}}_{filtered})$ 
            \vspace{5pt}
            \State \Return $\mathcal{P}_{filtered}, \: \mathcal{Q}_{\:filtered}$
        \EndFunction
    \end{algorithmic}
\end{algorithm}

The idea of our Octree Correspondence Filter is that if $\mathcal{P}$ and $\mathcal{Q}$ are stored in an octree data structure and are aligned perfectly, the occupancy of each voxel must be logically equivalent. 
Dynamic objects and non-overlapping areas, however, cause exclusive occupancy.
Points in these voxels can be filtered out.
Figure \ref{fig:octreeFilter} illustrates this for a moving car that causes exclusive occupancy, whereas a tree and a traffic sign have equivalent occupancy.

In detail, we implemented our Octree Correspondence Filter in the following way.
The first step is to find an appropriate transformation estimate $\mathbf{T}_{e}$ to align $\mathcal{P}$ with $\mathcal{Q}$. 
The exact transformation $\mathbf{T}$ will then be computed by the subsequent ICP procedure.
Due to the steady kinematics of an automobile, the transformation $\mathbf{T}$ performed by the car in the latest time frame can be approximated with the transformation $\mathbf{T}_{e, n-1}$ performed in the previous time frame.
This transformation estimation $\mathbf{T}_{e} = \mathbf{T}_{e, n-1}$ is then used to align the latest scanned point cloud $\mathcal{Q}$ and the previous one $\mathcal{P}$ in the global frame defined by $\mathcal{Q}$.
The next step is to convert the aligned point clouds $\mathcal{Q}$ and $\tilde{\mathcal{P}}=\mathbf{T}_{e}(\mathcal{P})$ into octree data structures $O_{\tilde{\mathcal{P}}}$ and $O_\mathcal{Q}$. 
This enables us to address the same individual voxel in both point clouds to compare them. 
This means that there are always two voxels $v_i \in O_{\tilde{\mathcal{P}}}$ and $u_i \in O_\mathcal{Q}$ referring to the same segment in space corresponding to the two point clouds. 
The algorithm then iterates through all occupied voxels in the global frame and checks if both point clouds have points located in their voxels $v_i \in O_{\tilde{\mathcal{P}}}$ and $u_j \in O_\mathcal{Q}$ at this particular position in space. 
If only one of the two voxels is occupied, these points likely do not possess any correct correspondences in the other point cloud and will be deleted. Last, the two octrees get converted back into point clouds $\mathcal{P}_{filtered}$ and $\mathcal{Q}_{\:filtered}$.

The algorithm requires few computational resources and, most importantly, scales well with a rising number of dynamic objects and points. 
In fact, it is even completely independent of the former, which would not be possible with explicit object detection and has an upper bound on the latter. 
The creation of the octree data structure is dependent on the size of the point cloud.
However, the iteration over the voxels is independent of the size of the point cloud and only depends on the depth of the octree.

\begin{table}
\begin{tabular}{llr}
\hline

\hline
data filtering         	     & edge points (\ref{sec:loamFeatures})\\
			     & planar points (\ref{sec:loamFeatures})\\
                             & Octree Correspondence Filter (\ref{sec:octreeFilter})\\
data processing              & $\mathbf{T}(\mathcal{P})$ \\       	
correspondence determination & closest point (\ref{icp_variants})\\
			     & reciprocal correspondence search (\ref{sec:reciprocal})    \\
			     & eigenvalues and eigenvectors (\ref{sec:edgePointsToEdgeLines} \& \ref{sec:planarPointToPlanarPatch}) \\
objective function           & point-to-point (\ref{sec:point-to-point})\\
      			     & point-to-plane (\ref{sec:point-to-plane})\\
      			     & symmetric objective function (\ref{sec:symmetric})     \\
			     & edge point to edge line (\ref{sec:edgePointsToEdgeLines}) \\
			     & planar point to planar patch (\ref{sec:planarPointToPlanarPatch}) \\
\hline
\end{tabular}
\caption{Summary of the ICP variants discussed in this paper and mapping of the ICP stages.}
\label{table:icp_variants}
\end{table}

\newpage
\section{Our Visualization Method}
\label{visualization_method}
\begin{figure}[t]
    \centering
    \includegraphics[width=0.7\textwidth]{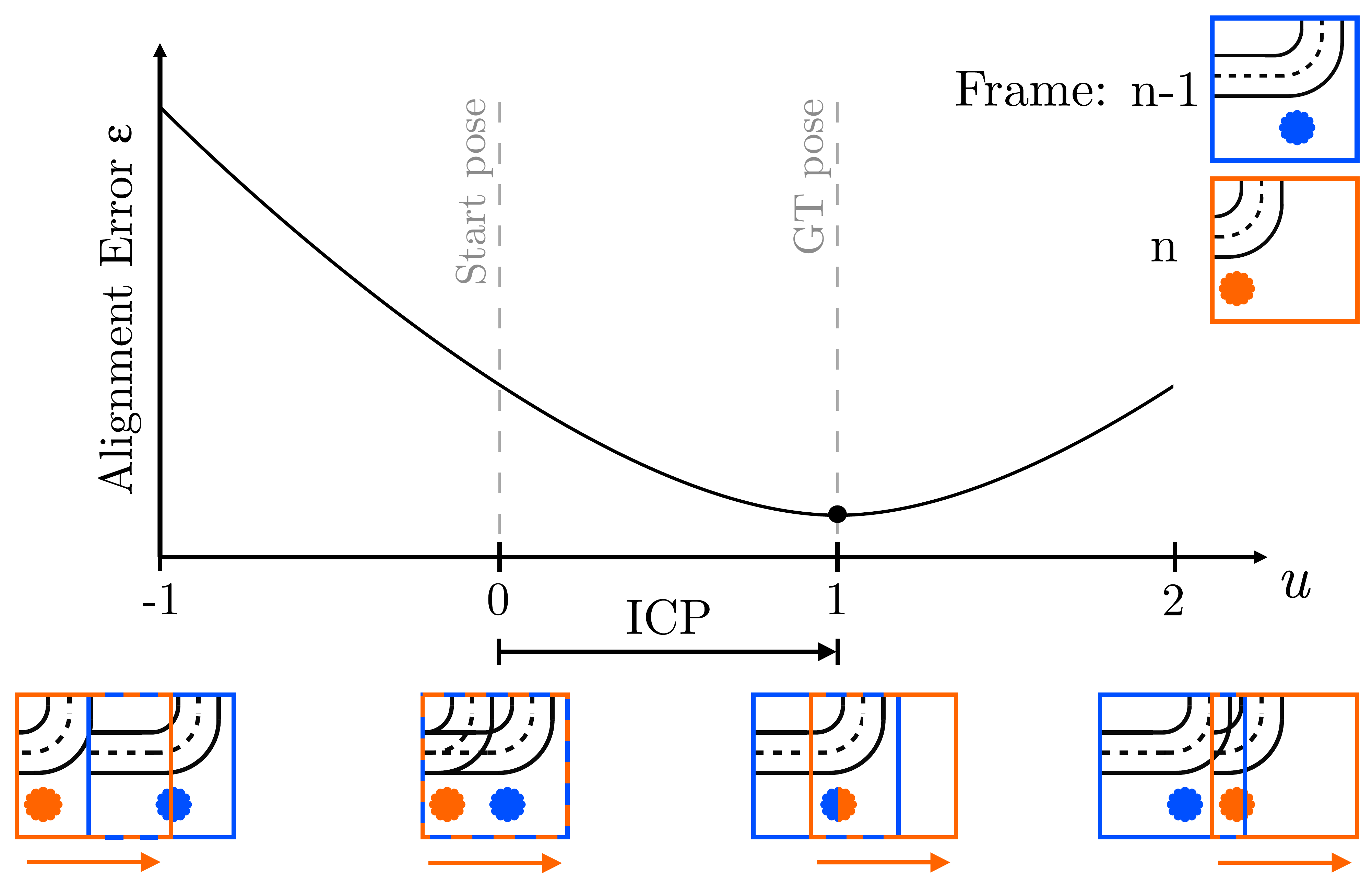}
\caption{The diagram shows two consecutive lidar scans, acquired from a car driving towards a left turn with a tree alongside the road. The most recent scan $n$, marked in orange, needs to be aligned within the reference coordinate system of the previous scan $n-1$, marked in blue. The start and end positions of the trajectory are at $u=0$ and $u=1$. The intervals $ [-1, 0) $ and $ (1, 2] $ evaluate the objective function for a finite set of extrapolated poses. The graph shows an ideal objective function, as the global minimum is at the ground truth pose at $ u=1 $.}
    \label{fig:Illustration of method}
\end{figure}

The previous sections have covered a variety of ICP approaches (cf. Table \ref{table:icp_variants}).
Although this list is not complete, it is already laborious to assess which variant to choose for lidar odometry.
One reason is that the performance of ICP variants is often measured on simulated datasets.
For lidar odometry, real-world data is more relevant since dynamic objects, non-overlapping areas, or sensor noise have a tremendous impact on the objective function.
The underlying reason is that those real-world effects cause local minima or an offset of the global minimum.
Therefore, we found it helpful to visualize the objective function.
A visualization shows the existence of multiple local minima and the location of the global minimum.
In conjunction with selected scenes (e.g. dynamic objects vs. no dynamic objects in the scene), this allows reviewing ICP pipelines in a qualitative way.

Unfortunately, it is inconvenient to visualize the objective function for lidar odometry.
Recalling the objective function $\epsilon(\mathbf{T})$, the mapping is $SE(3) \to \mathbb{R}$.
Although the codomain of the objective function is one-dimensional, the domain has six dimensions, which cannot be depicted in a single diagram.
In our conference paper, we have shown that for lidar odometry it is not necessary to vary the entire domain.
The main argument is that only a small subset is of interest.
If ICP converges towards the correct transformation $\mathbf{T}_{gt}$, the intermediate transformation estimates, e.g. $\text{ICP(i)} = \{ \mathbf{T}_0,\mathbf{T}_1,\mathbf{T}_2, ...\}$, usually lie on a direct path towards the ground truth $\mathbf{T}_{gt}$. 
If we evaluate the objective function for those intermediate transformations, e.g. $\{ \epsilon(\mathbf{T}_0),\epsilon(\mathbf{T}_1),\epsilon(\mathbf{T}_2), ...\}$, and use the iteration index $i \in \mathbb{N}$ as the domain, we can create a meaningful two-dimensional visualization since the mapping becomes $\epsilon(\text{ICP(i)}): \mathbb{N} \to \mathbb{R}$.
Although this approach is an option, there are some drawbacks. 
First, due to local minima, it is not guaranteed that ICP converges to the ground truth transformation $\mathbf{T}_{gt}$.
Second, we cannot control the distribution of the intermediate transformations because they are predetermined by ICP.
Third, comparing two different objective functions would not be possible because each ICP variant has its specific subset.
Since ICP takes an approximately shortest path, the idea of our conference paper is to interpolate between $\mathbf{T}_0$ and $\mathbf{T}_{gt}$.
This enables us to form an ICP-independent and controlled set of transformations $\text{INTERP(u)}=\{\mathbf{T}_{u}\}$.
Then we can use this set to visualize the objective function in two dimensions because the mapping is formally $\epsilon(\text{INTERP(u)}): \mathbb{R} \to \mathbb{R}$.
However, this approach has the drawback that an actual transformation estimate from ICP most often is not part of the interpolated set and therefore cannot be visualized in the same diagram.

Another explanation of our method is to view ICP as a closed control loop as depicted in Figure \ref{fig:icp_loop}. 
In the regular operation of ICP, each iteration computes a new transformation estimate $\mathbf{T}_e$.  
Our idea is to do an open-loop investigation by taking out the minimizer and using interpolated transformations $\mathbf{T}_u$.
This allows us to control the ICP pipeline and to query with specific transformations $\mathbf{T}_u$.
Expressing the interpolated transformations with the interpolation parameter $u$ makes it convenient to address certain transformations.

Figure \ref{fig:Illustration of method} illustrates our method.
As an example, a point cloud is moved along one axis.
The graph illustrates the objective function, whereas $u=0$ denotes the initial transformation and for $u=1$ the two point clouds are perfectly aligned.
For the interval $[0,1]$, we interpolate an arbitrary number of transformations.
Values outside of $[0,1]$ are extrapolated transformations.
We decided to interpolate linearly between $\mathbf{T}_0$ and $\mathbf{T}_{gt}$ to obtain a homogeneous distribution of interpolated transformations.
It is worth mentioning that for the interpolation, motion models such as the bicycle model could be used.

\subsection{Linear interpolation in $SE(3)$}
Linear interpolation of transformations in $SE(3)$ facilitates equally spaced transformations.
We treat the translation and the rotation part separately. 
With this simplification, we use linear interpolation (LERP) for the translation part and spherical linear interpolation (SLERP) for the rotation part.

Let $\mathbf{t}_i$ and $\mathbf{t}_j$ be the translation part of two consecutive transformations, and let $u$ be the interpolation parameter with $u \in [ 0 , 1 ]$, the LERP formula is then
\begin{equation}
\text{LERP} (\mathbf{t}_i, \mathbf{t}_j,u) = (1-u) \, \mathbf{t}_i + u \, \mathbf{t}_j \ .
\end{equation}
Analogous to LERP, a rotation can also be expressed by a linear combination of two rotations. 
We use the quaternion interpolation, introduced by Ken Shoemake \cite{Shoemake1985}. 
The advantage of SLERP is that it allows interpolation at constant rotational speed.

Let $\mathbf{q}_i$ and $\mathbf{q}_j$ be the quaternion representation of the rotation matrix $\mathbf{R}_i$ and $\mathbf{R}_j$, and let $u$ be the interpolation parameter, the SLERP formula is then
\begin{equation}
\text{SLERP} (\mathbf{q}_i,\mathbf{q}_j,u) = \mathbf{q}_i (\mathbf{q}^{-1}_i \mathbf{q}_j)^u  .
\end{equation}
Since we are using $u$ for LERP and SLERP, we are able to compute transformations between $\mathbf{T}_i$ and $\mathbf{T}_j$, and we will denote them with $\mathbf{T}_u$. Additionally, it is possible to compute extrapolated transformations by using values for $u$ outside the interval $[0,1]$.

\section{Experiments}
\label{experiments}
In order to demonstrate the power of our method, this section presents some experiments on real-world data.
The experiments were implemented in C++ using the Point Cloud Library (PCL) \cite{Rusu2011}.
The variety of possible experiments is very large.
Therefore, we conducted new experiments compared to those found in our conference paper.  
In this section, we structure our experiments into urban and highway scenarios to emphasize their differences. 
Then, we focus on experiments on highway scenes because they are more challenging.
Furthermore, to highlight the effect of dynamic objects we conduct experiments with and without dynamic objects.
For the plots, we normalize the objective function $\epsilon$ with the root-mean-square error (RMSE) to account for the distinct number of point correspondences of each variant.

\subsection{KITTI dataset}
As a representative dataset, we have chosen the well-known KITTI dataset \cite{Geiger2013}. 
Although it is relatively old, to the best of our knowledge, it is still the foremost dataset in terms of variety and size. 
The lidar is a Velodyne HDL64 mounted on the roof of a car. 
The odometry benchmark provides 11 sequences with ground truth data from an inertial navigation system (OXTS RT 3003). 
One advantage of this dataset is that the lidar scans are already de-skewed with the inertial navigation system. 
As a side note, as other researchers have found and as described in our conference paper, the ground truth data are sometimes several meters off \cite{Deschaud2018}. 
For this reason, we have carefully chosen sequences and frames with precise ground truth data.

\subsection{Parameters and modifications}
The ICP variants that we discussed in Section \ref{icp_variants} contain parameters that we want to disclose here. 
For the LOAM features from Section \ref{sec:loamFeatures}, we chose the following parameters. 
The planar points are classified as such if the $s_i$ value is smaller than $0.1$. 
All other points are regarded as edge points. 
For the computation of the eigenvalues and eigenvectors, a set of five closest points is used. 
An edge line is defined if the first eigenvalue is three times larger than the next smaller eigenvalue.
We regard points as a planar patch if the first and second-largest eigenvalues are three times larger than the smallest eigenvalue.
Furthermore, the patch should be flat.
Therefore, we used only planar patches where the smallest eigenvalue is smaller than $0.0001$.
Moreover, we subtract $\epsilon_{\text{planar-to-planarPatch}}$ with one, since the metric becomes one if the planar point matches perfectly the planar patch.

For the ego blind spot filter from Section \ref{sec:egoBlindSpot}, we encountered that the ego blind spot in the KITTI dataset is not perfectly circular.
Therefore, we create an artificial circular ego blind spot by removing points around the origin which fall in a circular area of radius $r=5\:m$.
After that, we applied the ego blind spot filter with radius $r=5\:m$.

For the Octree Correspondence Filter from Section \ref{sec:octreeFilter} an appropriate voxel size needs to be chosen. 
We oriented ourselves at the typical maximum deceleration of a modern car while emergency braking of \, $a_{brake} = -10\:m/s^2$ and, assuming it to be constant within the $\Delta \: t = 100\:ms$ time frame required to perform a full 360$^{\circ}$ lidar scan, the offset
\begin{equation}
	\Delta \: s(t) = \, s_1 - s_2 \, = \int v_0 \;\, dt - \int v_0 \; dt - \iint a_{brake} \;\, dt \, dt \;
\end{equation}
between the estimated location that does not anticipate the emergency braking $s_1$ and the actual position $s_2$ of the car after $t_1 = 0.1\:s$ is then
\begin{equation}
	\Delta \: s(t_1) \, = - 0.5 \; a_{brake} \; t_1^2 \; = \, 0.05 \; m \;.
\end{equation}
Given the voxel size of $0.1\:m$ used in our experiments and the worst-case offset of the estimated and actual car location being $0.05\:m$, the resulting misalignment of the two scans is, therefore, small enough to be negligible.
For the interpolation parameter $u$, we interpolate between $[-1,2]$ with 100 values.

\subsection{Static urban scenarios}
In this section, we evaluate the ICP variants for an urban environment with no dynamic objects in the scene. 
Moreover, we evaluate the difference between linear motion and rotational movement in turn situations. 

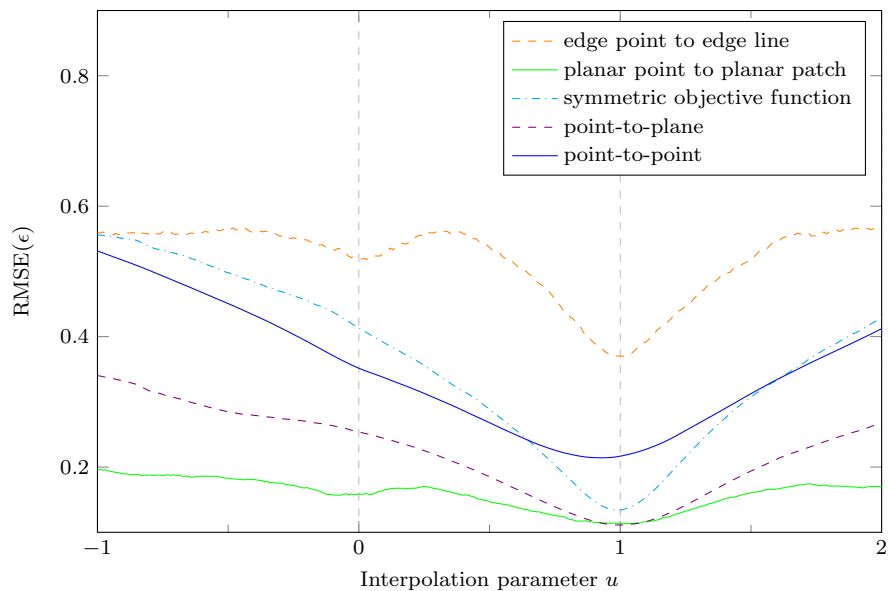
\begin{figure}
        \begin{tikzpicture}
            \begin{axis}[
            	width=119mm,
		height=85mm,
                ylabel={RMSE($ \epsilon $)},
                xlabel={Interpolation parameter $ u $},
                xmin=-1, xmax=2,
		ymin=0.1, ymax=0.9,
                xmajorgrids=true,
                ymajorgrids=false,
                xtick={-1,0, 1, 2},
                minor xtick={-0.5, 0.5, 1.5},
                major grid style={dashed},
                ylabel shift = 4 pt,
                legend cell align=left,
            ]
            
            \addplot[color=orange, style=dashed]
                table{data/0_1500_EDGEEDGELINESEGMENT_rmse.txt};
                \addlegendentry{edge point to edge line}
            \addplot[color=green]
                table{data/0_1500_PLANARPLANAR_rmse.txt};
                \addlegendentry{planar point to planar patch}
            \addplot[color=cyan, style=dashdotted]
                table{data/0_1500_SYMM_rmse.txt};
                \addlegendentry{symmetric objective function}
            \addplot[color=violet, style=dashed]
                table{data/0_1500_POINT_PLANE_rmse.txt};
                \addlegendentry{point-to-plane}
            \addplot[color=blue]
                table{data/0_1500_POINT_POINT_rmse.txt};
                \addlegendentry{point-to-point}

            \end{axis}
        \end{tikzpicture}
        \caption{KITTI (seq.: 00, frame: 1500): Overview of the performance of the objective functions on an urban scene with mainly straight movement. Except for the point-to-point metric, all objective functions have their global minima nearly at the ground truth transformation $\mathbf{T}_{gt}$.}
        \label{fig:00_1500}
	\end{figure}

Figure \ref{fig:00_1500} shows all objective functions evaluated for a frame where the vehicle's motion is mainly dominated by linear motion in the driving direction. 
The plane-related metrics, namely point-to-plane and symmetric objective function, have global minima at $u \approx 1$. 
Despite this, the gradient for the symmetric objective function is sharper compared to the point-to-plane metric. 

The feature-based metrics, edge point to edge line and planar point to planar patch, have their global minima close to $u=1$, too. 
Interestingly, they both have a local minimum around $u=0$. 
This is a hint that they are not robust against the non-overlapping area that is caused by the ego blind spot. 
The global minimum of the point-to-point metric has a slight offset to $u=1$. 
We experienced the same for other similar scenes. 
In conclusion, point-to-point is robust against local minima, but its global minimum does have an offset from the correct solution.

When vehicles conduct turns at intersections, the motion is dominated by rotational movement. 
Estimating accurate rotational movement is essential to detect loop closing. 
Inaccurately estimated angles have tremendous effects because they lever the estimated trajectory of the vehicle. 
For this reason, we analyzed the objective functions for frames when the vehicle is conducting a turn.

\begin{figure}
        \begin{tikzpicture}
            \begin{axis}[
            	width=119mm,
		height=85mm,
                ylabel={RMSE($ \epsilon $)},
                xlabel={Interpolation parameter $ u $},
                xmin=-1, xmax=2,
                xmajorgrids=true,
                ymajorgrids=false,
                xtick={-1,0, 1, 2},
                minor xtick={-0.5, 0.5, 1.5},
                major grid style={dashed},
                ylabel shift = 4 pt,
                legend cell align=left,
            ]
            
            \addplot[color=orange, style=dashed]
                table{data/0_100_EDGEEDGELINESEGMENT_rmse.txt};
                \addlegendentry{edge point to edge line}
            \addplot[color=green]
                table{data/0_100_PLANARPLANAR_rmse.txt};
                \addlegendentry{planar point to planar patch}
            \addplot[color=cyan, style=dashdotted]
                table{data/0_100_SYMM_rmse.txt};
                \addlegendentry{symmetric objective function}
            \addplot[color=violet, style=dashed]
                table{data/0_100_POINT_PLANE_rmse.txt};
                \addlegendentry{point-to-plane}
            \addplot[color=blue]
                table{data/0_100_POINT_POINT_rmse.txt};
                \addlegendentry{point-to-point}

            \end{axis}
        \end{tikzpicture}
        \caption{KITTI (seq.: 00, frame: 100): Overview of the performance of the objective functions in a turn scene with mainly rotational movement. All objective functions perform well and have the global minimum close to the ground truth.}
        \label{fig:00_100}
	\end{figure}
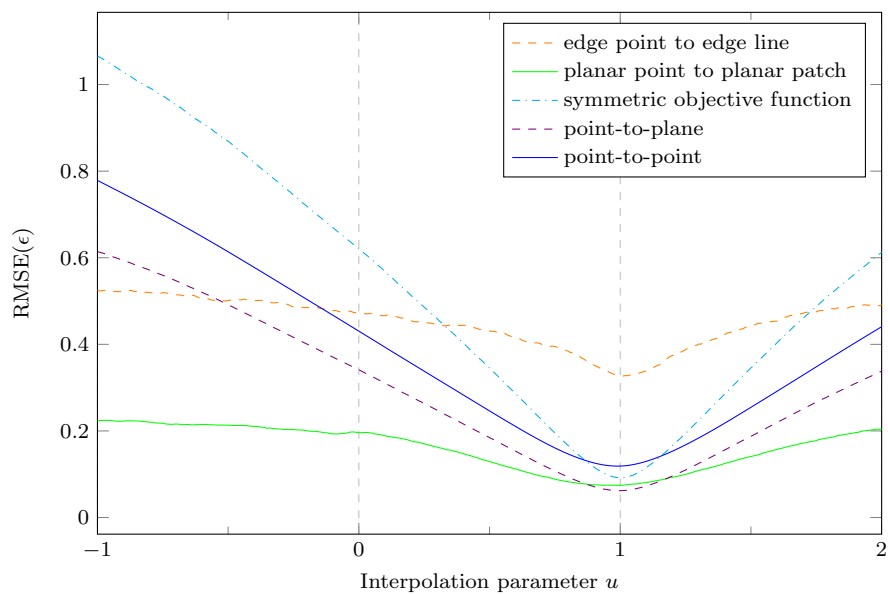

Figure \ref{fig:00_100} showcases the objective functions for a right turn with nearly no translational movement. 
All objective functions have their global minimum close to $u = 1$. 
For other turn scenes, we have encountered the same, and we conclude that to estimate rotational movement in urban environments, the choice of the objective function is less important because all work well. 

\subsection{Highway scenarios}
Highway scenarios are, in general, more challenging for lidar odometry than urban scenarios because of the lack of buildings alongside the road. 
Occasional structures like tree trunks or poles from traffic lights are, therefore, the only landmarks that can be used to estimate movement. 
Unfortunately, often only a few lidar measurements are obtained from these thin structures. 
Another challenge originates from other traffic participants that often have large relative velocity differences.
To study the effect of dynamic objects independently from the static highway environment, we first conducted experiments without other vehicles around the ego vehicle. 
Afterwards, we investigated the effect of a vehicle that is passing the ego vehicle close by. 

\subsubsection{No dynamic objects}
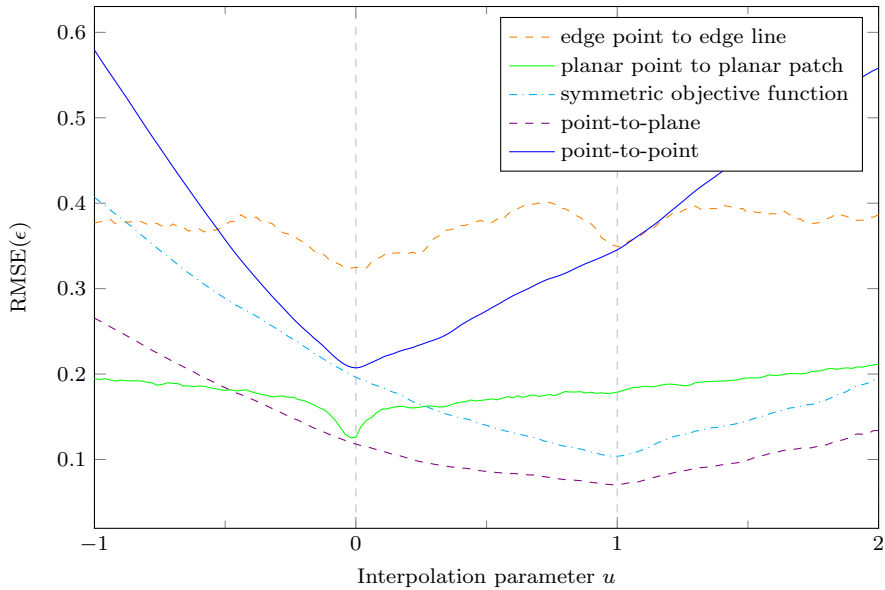
\begin{figure}
        \begin{tikzpicture}
            \begin{axis}[
            	width=119mm,
		height=85mm,
                ylabel={RMSE($ \epsilon $)},
                xlabel={Interpolation parameter $ u $},
                xmin=-1, xmax=2,
                xmajorgrids=true,
                ymajorgrids=false,
                xtick={-1,0, 1, 2},
                minor xtick={-0.5, 0.5, 1.5},
                major grid style={dashed},
                ylabel shift = 4 pt,
                legend cell align=left,
            ]
            
            \addplot[color=orange, style=dashed]
                table{data/1_320_EDGEEDGELINESEGMENT_rmse.txt};
                \addlegendentry{edge point to edge line}
            \addplot[color=green]
                table{data/1_320_PLANARPLANAR_rmse.txt};
                \addlegendentry{planar point to planar patch}
            \addplot[color=cyan, style=dashdotted]
                table{data/1_320_SYMM_rmse.txt};
                \addlegendentry{symmetric objective function}
            \addplot[color=violet, style=dashed]
                table{data/1_320_POINT_PLANE_rmse.txt};
                \addlegendentry{point-to-plane}
            \addplot[color=blue]
                table{data/1_320_POINT_POINT_rmse.txt};
                \addlegendentry{point-to-point}

            \end{axis}
        \end{tikzpicture}
        \caption{KITTI (seq.: 01, frame: 320): Highway scene with no other dynamic objects nearby. The point-to-point and planar point to planar patch metrics struggle since their global minima are at the initial position. Although the edge point to edge line metric has a global minimum at $u \approx 0$, it has a local minimum at $u \approx 1$. The point-to-plane and symmetric objective function have their global minima at $u \approx 1$ and exhibit a large basin of convergence.}
        \label{fig:01_320}
	\end{figure}

Figure \ref{fig:01_320} provides an overview of the objective functions for a highway scene with no other dynamic objects nearby. 
The point-to-point and planar point to planar patch metrics do not perform well. 
They have their global minima at the initial position. 
We explain this with the high concentration of points around the vehicle that leads to more correspondences than at the correct position $u=1$. 
The point-to-plane and symmetric objective function use their property to slide more easily along the road surface and have their global minima at $u=1$. 
The edge point to edge line metric, however, has a unique behavior. 
Although the global minimum is at $u \approx 0$, it has a local minimum at $u \approx 1$.
We explain this with edge points that are not located at real edges but irregularity in the data leads to the computation of edge points. 

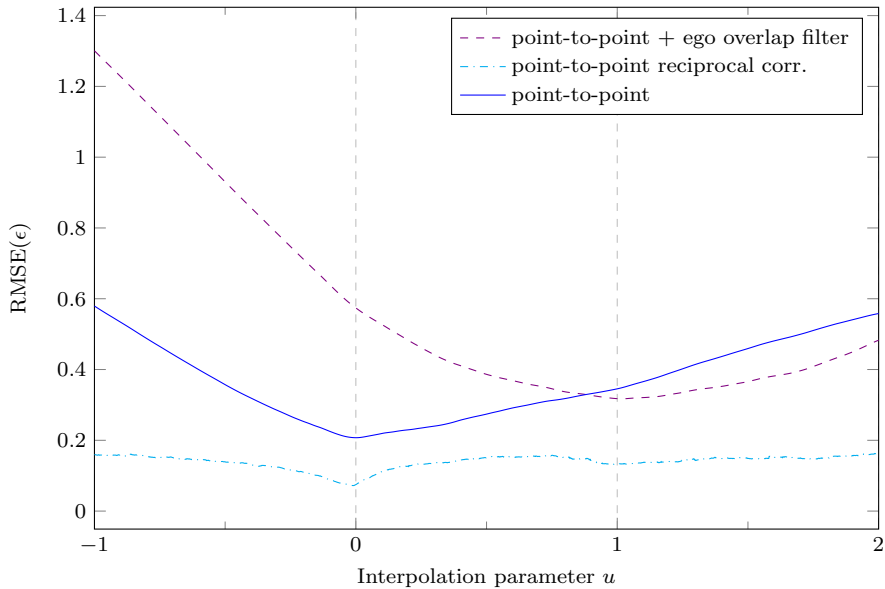
\begin{figure}
        \begin{tikzpicture}
            \begin{axis}[
            	width=119mm,
		height=85mm,
                ylabel={RMSE($ \epsilon $)},
                xlabel={Interpolation parameter $ u $},
                xmin=-1, xmax=2,
                xmajorgrids=true,
                ymajorgrids=false,
                xtick={-1,0, 1, 2},
                minor xtick={-0.5, 0.5, 1.5},
                major grid style={dashed},
                ylabel shift = 4 pt,
                legend cell align=left,
            ]
            
            \addplot[color=violet, style=dashed]
                table{data/1_320_POINT_POINT_offlineOverlap_6.txt};
                \addlegendentry{point-to-point + ego overlap filter}
            \addplot[color=cyan, style=dashdotted]
                table{data/1_320_POINT_POINT_reciprocal.txt};
                \addlegendentry{point-to-point reciprocal corr.}
            \addplot[color=blue]
                table{data/1_320_POINT_POINT_rmse.txt};
                \addlegendentry{point-to-point}

            \end{axis}
        \end{tikzpicture}
        \caption{KITTI (seq.: 01, frame: 320): Improvement of the point-to-point metric by applying the reciprocal correspondence method and the ego overlap filter. The point-to-point metric with the reciprocal correspondence method preserves its global minimum at $u \approx 0$, but the reciprocal correspondences cause a local minimum at $u \approx 1$. If the effect of the ego blind spot is removed by the ego overlap filter, the point-to-point metric has a global minimum at $u \approx 1$.}
        \label{fig:01_320_imp}
	\end{figure}

One reason for the poor performance of the point-to-point metric is the non-overlapping area around the vehicle.
The reciprocal correspondence method and the ego overlap filter should mitigate this problem. 
Figure \ref{fig:01_320_imp} presents the outcome when applying these preprocessing steps to the point-to-point metric. 
The reciprocal correspondence search slightly improves the performance of the point-to-point metric. 
If the transformation $\mathbf{T}_{e,n-1}$ from the previous frame is used to apply the ego overlap filter, then the point-to-point metric has its global minimum at $u \approx 1$.

\subsubsection{Dynamic objects}
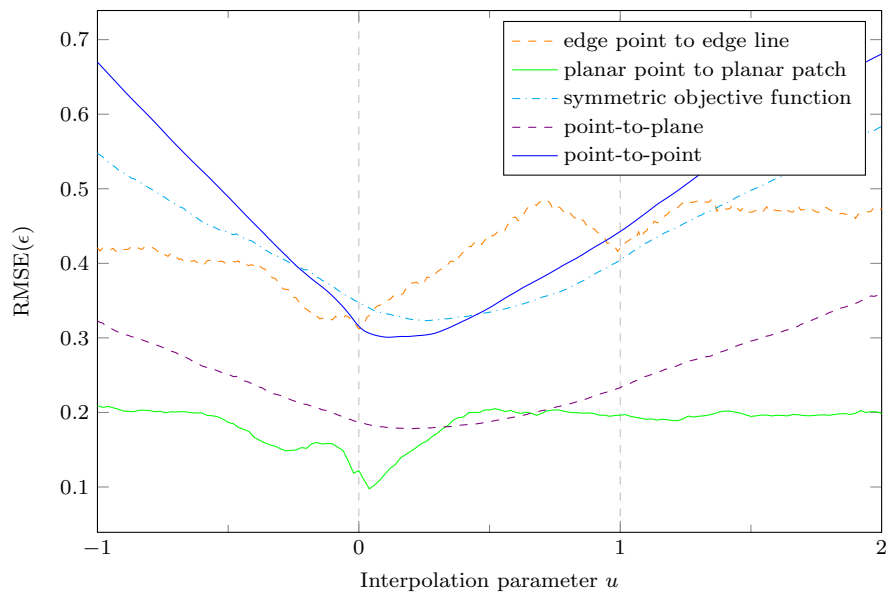
\begin{figure}
        \begin{tikzpicture}
            \begin{axis}[
            	width=119mm,
		height=85mm,
                ylabel={RMSE($ \epsilon $)},
                xlabel={Interpolation parameter $ u $},
                xmin=-1, xmax=2,
                xmajorgrids=true,
                ymajorgrids=false,
                xtick={-1,0, 1, 2},
                minor xtick={-0.5, 0.5, 1.5},
                major grid style={dashed},
                ylabel shift = 4 pt,
                legend cell align=left,
            ]
            
            \addplot[color=orange, style=dashed]
                table{data/1_420_EDGEEDGELINESEGMENT_edges.txt};
                \addlegendentry{edge point to edge line}
            \addplot[color=green]
                table{data/1_420_PLANARPLANAR.txt};
                \addlegendentry{planar point to planar patch}
            \addplot[color=cyan, style=dashdotted]
                table{data/1_420_SYMM.txt};
                \addlegendentry{symmetric objective function}
            \addplot[color=violet, style=dashed]
                table{data/1_420_POINT_PLANE.txt};
                \addlegendentry{point-to-plane}
            \addplot[color=blue]
                table{data/1_420_POINT_POINT.txt};
                \addlegendentry{point-to-point}

            \end{axis}
        \end{tikzpicture}
        \caption{KITTI (seq.: 01, frame: 420): Overview of the performance of the objective functions on a highway scene with other dynamic objects close by. The point-to-point and planar point to planar patch metrics struggle the most. Although the edge point to edge line metric has its global minimum at $u \approx 0$, it has a local minimum at $u \approx 1$.}
        \label{fig:01_420}
	\end{figure}

As previously mentioned, dynamic objects are the most challenging aspect of point cloud registration and lidar odometry. 
To highlight this fact, Figure \ref{fig:01_420} depicts the graphs of the objective functions for a highway scene with other dynamic objects close by. 
All objective functions have their global minimum around $u=0$. 
Only the edge point to edge line metric has a local minimum at $u \approx 1$. 
The reason for the poor performance of all other metrics is the presence of a vehicle close by which drives faster than the ego vehicle. 
Since the distribution of measurements is higher at the vicinity of the ego vehicle, a lot of wrong correspondences originate from this vehicle.

\begin{figure}
        \begin{tikzpicture}
            \begin{axis}[
            	width=119mm,
		height=85mm,
                ylabel={RMSE($ \epsilon $)},
                xlabel={Interpolation parameter $ u $},
                xmin=-1, xmax=2,
                xmajorgrids=true,
                ymajorgrids=false,
                xtick={-1,0, 1, 2},
                minor xtick={-0.5, 0.5, 1.5},
                major grid style={dashed},
                ylabel shift = 4 pt,
                legend cell align=left,
            ]
            
            \addplot[color=orange, style=dashed]
                table{data/1_420_EDGEEDGELINESEGMENT_ocf_rmse.txt};
                \addlegendentry{edge point to edge line + OCF}
            \addplot[color=green]
                table{data/1_420_PLANARPLANAR_ocf_rmse.txt};
                \addlegendentry{planar point to planar patch + OCF}
            \addplot[color=cyan, style=dashdotted]
                table{data/1_420_SYMM_ocf_rmse.txt};
                \addlegendentry{symmetric objective function + OCF}
            \addplot[color=violet, style=dashed]
                table{data/1_420_POINT_PLANE_ocf_rmse.txt};
                \addlegendentry{point-to-plane + OCF}
            \addplot[color=blue]
                table{data/1_420_POINT_POINT_ocf_rmse.txt};
                \addlegendentry{point-to-point + OCF}

            \end{axis}
        \end{tikzpicture}
        \caption{KITTI (seq.: 01, frame: 420): Improvement through the Octree Correspondence Filter. All objective functions have their global minima close to $u = 1$. The point-to-point metric possesses the highest gradient.}
        \label{fig:01_420_octree}
	\end{figure}
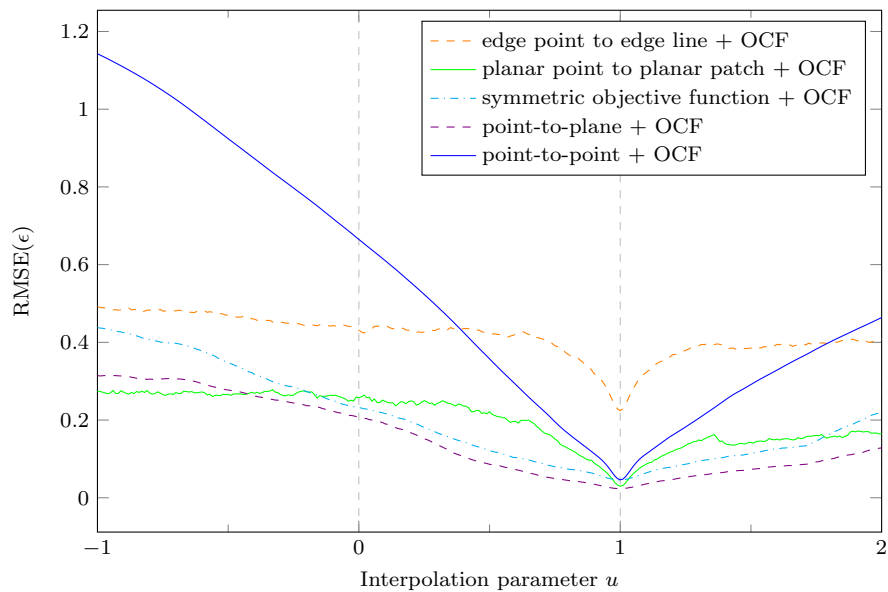

A data filter needs to be applied to exclude the points which belong to the dynamic objects. Figure \ref{fig:01_420_octree} presents the improvement when our Octree Correspondence Filter is applied to this scene. 
All objective functions have their global minimum at $u \approx 1$.
The point-to-point metric has a higher gradient than the point-to-plane and symmetric objective function.
In conclusion, the Octree Correspondence Filter is powerful in improving the performance of any metric.
This is clearly visible with our visualization method.

\section{Conclusion and Future Directions}
\label{conclusion}
In this article, we visually studied ICP variants in the context of lidar odometry. 
We clustered the ICP pipeline into five stages. 
The ICP pipeline can be seen as a control loop where the last stage, the minimizer, provides an estimate for the current transformation, which is fed back into the earlier stages. 
Our idea is to leave out the minimizer and to use interpolated transformations from ground truth data to analyze the pre- and processing stages of different ICP variants and to plot the objective function in two dimensions.

We demonstrated the capability of our method on the KITTI odometry benchmark with the following insights. 
First, we revisited our former work by analyzing linear motion on a straight road and during purely rotational motion in a turn situation.
Our visualization showed that for rotational motion, all objective functions performed well. 
For linear motion, the point-to-point metric has a small offset from the ground truth.
For the LOAM features, we showed that the edge point to edge line metric provides better global minima than the planar point to planar patch metric.
Moreover, the point-to-plane and symmetric objective function, which are conceptually similar to the planar point to planar patch metric, often perform better.

We think our visual method is a convenient tool to analyze ICP variants for certain applications and to develop new variants.
We have shown that non-overlapping areas in the point clouds have a strong influence on the objective function if they are not considered. 
In the case of lidar odometry, one source of non-overlapping areas is the ego blind spot. 
Since the shape of the blind spot is known beforehand, points that lie in this area can be filtered out.
Another example is the Octree Correspondence Filter that we developed. 
With our visualization, it becomes clearly visible that the Octree Correspondence Filter is able to minimize the effect of dynamic objects and non-overlapping areas. 

Our future work will focus on evaluating whether our visual method is applicable to probabilistic ICP methods such as G-ICP and NDT.
Moreover, the qualitative nature of our analysis might be extended in a quantitative way.


%
%

\bibliographystyle{spmpsci}      
\bibliography{bibliography}   

\begin{thebibliography}{10}
\providecommand{\url}[1]{{#1}}
\providecommand{\urlprefix}{URL }
\expandafter\ifx\csname urlstyle\endcsname\relax
  \providecommand{\doi}[1]{DOI~\discretionary{}{}{}#1}\else
  \providecommand{\doi}{DOI~\discretionary{}{}{}\begingroup
  \urlstyle{rm}\Url}\fi

\bibitem{Bengtsson1999}
Bengtsson, O., Baerveldt, A.J.: Localization in changing environments by
  matching laser range scans.
\newblock In: 1999 Third European Workshop on Advanced Mobile Robots
  (Eurobot{\textquotesingle}99). Proceedings (Cat. No.99EX355). {IEEE} (1975).
\newblock \doi{10.1109/eurbot.1999.827637}

\bibitem{Bentley1975}
Bentley, J.L.: Multidimensional binary search trees used for associative
  searching.
\newblock Commun. ACM \textbf{18}(9), 509–517 (1975).
\newblock \doi{10.1145/361002.361007}.
\newblock \urlprefix\url{https://doi.org/10.1145/361002.361007}

\bibitem{Besl1992}
{Besl}, P.J., {McKay}, N.D.: A method for registration of {3-D} shapes.
\newblock IEEE Transactions on Pattern Analysis and Machine Intelligence
  \textbf{14}(2), 239--256 (1992).
\newblock \doi{10.1109/34.121791}

\bibitem{Chen1991}
{Chen}, Y., {Medioni}, G.: Object modeling by registration of multiple range
  images.
\newblock In: Proceedings. 1991 IEEE International Conference on Robotics and
  Automation, pp. 2724--2729 vol.3 (1991).
\newblock \doi{10.1109/ROBOT.1991.132043}.
\newblock
  \urlprefix\url{https://graphics.stanford.edu/~smr/ICP/comparison/chen-medioni-align-rob91.pdf}

\bibitem{Deschaud2018}
Deschaud, J.: {IMLS-SLAM:} scan-to-model matching based on {3D} data.
\newblock CoRR \textbf{abs/1802.08633} (2018).
\newblock \urlprefix\url{http://arxiv.org/abs/1802.08633}

\bibitem{Dingler2019}
{Dingler}, S., {Burrichter}, H.: A visual method to study the error function of
  {ICP} algorithms.
\newblock In: 2019 19th International Conference on Advanced Robotics (ICAR),
  pp. 278--283 (2019).
\newblock \doi{10.1109/ICAR46387.2019.8981610}

\bibitem{Geiger2013}
Geiger, A., Lenz, P., Stiller, C., Urtasun, R.: Vision meets robotics: The
  {KITTI} dataset.
\newblock International Journal of Robotics Research (IJRR)  (2013).
\newblock \urlprefix\url{http://www.cvlibs.net/publications/Geiger2013IJRR.pdf}

\bibitem{Landry2019}
{Landry}, D., {Pomerleau}, F., {Giguère}, P.: Cello-{3D}: Estimating the
  covariance of {ICP} in the real world.
\newblock In: 2019 International Conference on Robotics and Automation (ICRA),
  pp. 8190--8196 (2019).
\newblock \doi{10.1109/ICRA.2019.8793516}

\bibitem{Magnusson2009b}
Magnusson, M.: The three-dimensional normal-distributions transform --- an
  efficient representation for registration, surface analysis, and loop
  detection.
\newblock Ph.D. thesis (2009)

\bibitem{Magnusson2009}
{Magnusson}, M., {Nuchter}, A., {Lorken}, C., {Lilienthal}, A.J., {Hertzberg},
  J.: Evaluation of {3D} registration reliability and speed - a comparison of
  {ICP} and {NDT}.
\newblock In: 2009 IEEE International Conference on Robotics and Automation,
  pp. 3907--3912 (2009).
\newblock \doi{10.1109/ROBOT.2009.5152538}

\bibitem{Mitra2004}
Mitra, N.J., Gelfand, N., Pottmann, H., Guibas, L.: Registration of point cloud
  data from a geometric optimization perspective.
\newblock In: Symposium on Geometry Processing, pp. 23--31 (2004)

\bibitem{Pajdla1995}
{Pajdla}, T., {Van Gool}, L.: Matching of {3-D} curves using semi-differential
  invariants.
\newblock In: Proceedings of IEEE International Conference on Computer Vision,
  pp. 390--395 (1995)

\bibitem{Pomerleau2013}
Pomerleau, F., Colas, F., Siegwart, R., Magnenat, S.: Comparing {ICP} variants
  on real-world data sets.
\newblock Autonomous Robots \textbf{34}(3), 133--148 (2013).
\newblock \doi{10.1007/s10514-013-9327-2}.
\newblock \urlprefix\url{https://doi.org/10.1007/s10514-013-9327-2}

\bibitem{Rusinkiewicz2019}
Rusinkiewicz, S.: A symmetric objective function for {ICP}.
\newblock ACM Transactions on Graphics (Proc. SIGGRAPH) \textbf{38}(4) (2019)

\bibitem{Rusinkiewicz2001}
{Rusinkiewicz}, S., {Levoy}, M.: Efficient variants of the {ICP} algorithm.
\newblock In: Proceedings Third International Conference on {3-D} Digital
  Imaging and Modeling, pp. 145--152 (2001).
\newblock \doi{10.1109/IM.2001.924423}

\bibitem{Rusu2011}
{Rusu}, R.B., {Cousins}, S.: {3D} is here: Point cloud library ({PCL}).
\newblock In: 2011 IEEE International Conference on Robotics and Automation,
  pp. 1--4 (2011).
\newblock \doi{10.1109/ICRA.2011.5980567}

\bibitem{Segal2009}
Segal, A., Hähnel, D., Thrun, S.: Generalized-{ICP} (2009).
\newblock \doi{10.15607/RSS.2009.V.021}

\bibitem{Shoemake1985}
Shoemake, K.: Animating rotation with quaternion curves.
\newblock In: Proceedings of the 12th Annual Conference on Computer Graphics
  and Interactive Techniques, SIGGRAPH '85, pp. 245--254. ACM, New York, NY,
  USA (1985).
\newblock \doi{10.1145/325334.325242}.
\newblock \urlprefix\url{http://doi.acm.org/10.1145/325334.325242}

\bibitem{Tazir2018}
Tazir, M.L., Gokhool, T., Checchin, P., Malaterre, L., Trassoudaine, L.:
  {CICP}: Cluster iterative closest point for sparse-dense point cloud
  registration.
\newblock Robotics and Autonomous Systems \textbf{108} (2018).
\newblock \doi{10.1016/j.robot.2018.07.003}

\bibitem{Zhang2014}
Zhang, J., Singh, S.: {LOAM}: Lidar odometry and mapping in real-time.
\newblock In: Robotics: Science and Systems (2014).
\newblock \doi{10.15607/RSS.2014.X.007}.
\newblock \urlprefix\url{http://www.roboticsproceedings.org/rss10/p07.pdf}

\end{thebibliography}

%
%

\end{document}